\documentclass[preprint,number,12pt]{elsarticle}
\usepackage{lineno}
\usepackage{fmtcount}
\usepackage[colorlinks,linkcolor=blue,urlcolor=blue]{hyperref}
\usepackage{graphicx,verbatim}
\usepackage{float}
\usepackage{amssymb}
\usepackage{amsmath}
\usepackage{amssymb}
\usepackage{multirow}
\usepackage{tabularx}
\usepackage{gensymb}
\usepackage{url}
\usepackage{makecell}
\usepackage[table,xcdraw]{xcolor}
\usepackage{amsmath}
\usepackage{xcolor}
\usepackage{graphicx} 
\usepackage{booktabs}
\usepackage{mathrsfs}
\usepackage{pifont}
  \newif\ifblind
\blindtrue

\usepackage{setspace}

\let\oldbibliography\thebibliography
\renewcommand{\thebibliography}[1]{%
  \oldbibliography{#1}%
  \setstretch{0.9}%
}

\begin{document}
\begin{frontmatter}

\title{Deep Learning-based Intelligent Diagnosis of Congenital Uterine Anomalies in 3D Ultrasound}
\author[1]{Yueyue Xu\fnref{fn1}}
    \author[2,3]{Yuhao Huang\fnref{fn1}}
    \fntext[fn1]{These authors contributed equally to this work.}
    \author[2]{Jiaxiao Deng}
    \author[2,4]{Yuanji Zhang}
    \author[2]{Haoming Zhang}
    \author[1]{Jiajia Qu}
    \author[1]{Shiying Zheng}
    \author[1]{Xiaomei Tang}
    \author[1]{Haining Chen}
    \author[5]{Chengcai Chen}
    \author[6]{Yiyi Wu}
    \author[7]{Xin Yang}
    \author[2,8,9]{Dong Ni}
    \author[1]{Hongyu Zheng\corref{cor1}}
    \ead{187217966@qq.com}
    \cortext[cor1]{Corresponding author.}
    \affiliation[1]{organization={The People’s Hospital of Guangxi Zhuang Autonomous Region}, city={Nanning}, state={Guangxi}, country={China}}
    \affiliation[2]{organization={Medical Ultrasound Image Computing (MUSIC) Lab, Shenzhen University}, city={Shenzhen}, state={Guangdong}, country={China}}
    \affiliation[3]{organization={Centre for Artificial Intelligence and Robotics, Hong Kong Institute of Science \& Innovation, Chinese Academy of Sciences}, city={Hong Kong}, country={China}}
    \affiliation[4]{organization={Shenzhen Luohu People's Hospital (The Third Affiliated Hospital of Shenzhen University)}, city={Shenzhen}, state={Guangdong}, country={China}}
    \affiliation[5]{organization={Affiliated Hospital of Youjiang Medical College for Nationalities}, city={Youjiang}, state={Guangxi}, country={China}}
    \affiliation[6]{organization={Guilin Maternal and Child Health Hospital}, city={Guilin}, state={Guangxi}, country={China}}
    \affiliation[7]{organization={School of Biomedical Engineering, Medical School, Shenzhen University}, city={Shenzhen}, state={Guangdong}, country={China}}
    \affiliation[8]{organization={School of Artificial Intelligence, Shenzhen University}, city={Shenzhen}, state={Guangdong}, country={China}}
    \affiliation[9]{organization={School of Biomedical Engineering and Informatics, Nanjing Medical University}, city={Nanjing}, state={Jiangsu}, country={China}}

\begin{abstract}
\textit{Objective:} To develop an intelligent framework, termed CUA-Net, for the automated classification of congenital uterine anomalies (CUA) without requiring coronal plane reconstruction, and to evaluate its clinical applicability.\\
\textit{Methods:} CUA-Net was built on 3D ResNet-18, equipped with a dynamic data resampling strategy to mitigate the data imbalance issue and a hard sample mining technique to fully learn from the difficult cases by loss adjustment.
We further proposed the self-supervised reconstruction to comprehensively explore the volumes and the online data augmentation to refine the wrong predictions and enhance the model’s generalization.
We compared the CUA-Net with different deep-learning methods and junior/senior sonographers in the testing set.
The evaluation metrics included accuracy, precision, recall, F1-score, micro-AUC, and macro-AUC.\\
\textit{Results:} The proposed CUA-Net exhibited satisfactory performance in both internal and external test sets.
In the internal cohort, the model achieved accuracy of 93.88\%, precision of 87.01\%, recall of 95.92\%, F1-score of 88.09\%, and micro-AUC of 0.9982 and macro-AUC of 0.9997.
In the external set, it maintained good performance with accuracy of 91.52\%, precision of 83.27\%, recall of 88.63\%, F1-score of 81.49\%, micro-AUC of 0.9945 and macro-AUC of 0.9990.
Our CUA-Net outperformed the junior sonographers across all performance indicators and achieved performance comparable to that of the senior sonographers across most metrics.\\
\textit{Conclusion:} 
The CUA-Net demonstrates favorable accuracy and generalizability in classifying common CUA categories, while showing preliminary potential for recognizing less prevalent anomalies. 
These capabilities may help optimize clinical workflows and support more standardized diagnosis.
\end{abstract}

\begin{keyword}
Congenital Uterine Anomalies, 3D Ultrasound, Deep Learning, Classification, Diagnostic Workflow
\end{keyword}
\end{frontmatter}

\section{Introduction}
\label{sec1}

Congenital Uterine Anomalies (CUA) refer to anatomical abnormalities caused by incomplete fusion or resorption of the paramesonephric (Müllerian) ducts during embryonic development~\cite{bortoletto2024mullerian}. CUA is one of the leading causes of female infertility, recurrent miscarriage, intrauterine growth restriction, preterm birth, and retained placenta. Studies have shown that its overall prevalence is estimated to be 5.5–6.7\% in the general population, 7.3–8.0\% among infertile women, and as high as 16.7\% in those with recurrent spontaneous abortions~\cite{dietrich2022diagnosis,abhinaya2024study}. 
There are various types of CUAs, including septate uterus, arcuate uterus, uterus didelphys, unicornuate uterus, T-shaped uterus,  and bicornuate uterus~\cite{am1988american}.
In clinical practice, the prevalence of different CUA subtypes is highly imbalanced. 
According to a systematic review by Chan et al., in unselected populations, the reported prevalence is 3.9\% for arcuate uterus, 2.3\% for septate uterus, and 0.4\% for bicornuate uterus~\cite{chan2011prevalence}.
For T-shaped uterus, although reported prevalence varies because of inconsistent diagnostic criteria, a recent prospective study using the CUME criteria reported a prevalence of only 0.8\% among fertile women~\cite{coelho2021definition,seyhan2021prevalence}.
Since different types of CUA require distinct clinical interventions, and some forms can significantly improve fertility and pregnancy outcomes after surgical correction, accurate diagnosis and classification of uterine anomalies are of great importance for guiding clinical treatment decisions~\cite{e2020diagnosis}.

\begin{figure}[!t]
	\centering
	\includegraphics[width=1.0\linewidth]{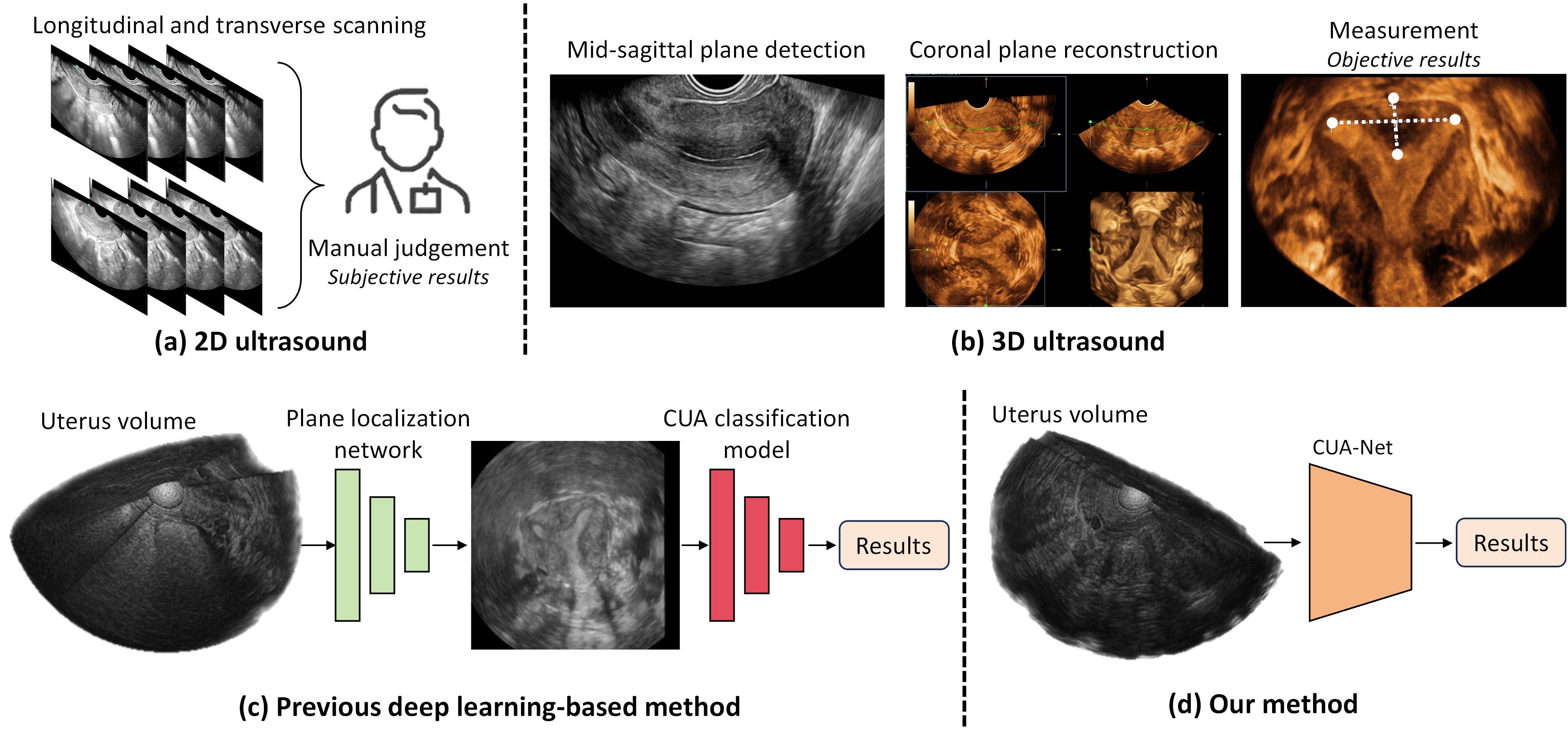}
	\caption{Comparison of different workflows for CUA classification.}
	\label{fig:intro}
\end{figure}

Ultrasound (US) serves as a primary auxiliary tool in the diagnosis and classification of CUA due to its distinct advantages over other modalities, e.g., hysterosalpingography, magnetic resonance imaging, hysteroscopy, and laparoscopy~\cite{ludwin2019reproductive}. 
It is a non-invasive, radiation-free technique that offers real-time imaging, making it particularly suitable for dynamic assessment and routine use in obstetrics and gynecology~\cite{huang2020searching,zhang2026artificial}.
US also allows observation of the endometrial cavity and the outer contour of the uterus through transverse and longitudinal sections.
However, conventional 2D scanning cannot capture the coronal plane, which limits its ability to distinguish between different types of uterine anomalies.
It highly relies on sonographers’ experience to output the subjective results based on longitudinal and transverse scanning (Fig.~\ref{fig:intro} (a)).
Currently, a definitive diagnosis often requires further invasive procedures such as hysteroscopy or laparoscopy.

In contrast, 3D US enables visualization of the uterine coronal plane, providing a comprehensive view of the uterine structure and its spatial relationships, thereby supporting accurate diagnosis and classification of CUAs.
However, 3D acquisitions of uterine coronal planes rely on high-quality images of the midsagittal plane, followed by a coronal sweep that allows for a good reconstruction of the uterine volume.
Sonographers must be highly skilled with years of clinical experience to perform this technique.
Accurate classification of uterine anomalies also typically involves detailed measurements on the obtained coronal plane (Fig.~\ref{fig:intro} (b)).
Furthermore, constrained by the known disparities in incidence rates among subtypes, the diagnosis and identification of certain rare congenital uterine anomalies remain particularly challenging.
These challenges may increase the complexity and time required for diagnosis, often leading to low consistency and accuracy, especially in complex anomaly cases.

In recent years, artificial intelligence and deep learning (DL) have made significant advances in the field of medical image analysis~\cite{shen2017deep,huang2023fourier,wang2025deep,zhou2026ctrl}, especially for intelligent US~\cite{huang2024segment,huang2025flip,chen2025enhancing,liu2025mreg,liang2026faa}.
Researchers studied the automatic methods for uterine fibroid detection~\cite{yang2023real} and classification~\cite{dilna2022classification} in US images.
Boneš et al. developed an automated system for the segmentation and alignment of uterine shapes from 3D US~\cite{bonevs2024automatic}.
Yang et al. first proposed reinforcement learning-based methods to locate the uterine coronal plane in 3D US~\cite{yang2021agent,yang2021searching}.
Following the above framework, Zou et al. further enhanced the model flexibility and accuracy in detecting uterine planes~\cite{zou2022agent}.
The above methods are capable of iteratively extracting the uterine coronal plane and hold potential for enabling CUA diagnosis in subsequent steps~\cite{huang2023localizing}. 
However, accurate CUA analysis relies on measuring multiple parameters on the coronal plane, which requires additional annotations such as keypoints, measurement lines, or segmentations. 
Furthermore, accumulated errors of plane localization ($>$10\degree~spatial angular error in~\cite{zou2022agent}) can significantly compromise the accuracy of downstream CUA assessment.

Most recently, Dou et al.~\cite{dou2025standard} introduced a diffusion-based approach to handle the plane localization task in 3D uterine volumes.
They also explored an uncertainty-driven strategy for binary classification of normal versus CUA cases, achieving an accuracy of 90.67\% (Fig.~\ref{fig:intro} (c)).
However, their approaches still rely on dynamic modeling of the plane localization process.
At present, there is a lack of intelligent methods that directly analyze 3D data to achieve fine-grained and multi-class CUA classification.

In this work, we proposed a deep learning-based framework for automatic CUA classification in 3D US, named CUA-Net.
For 3D uterine volumes, it can automatically distinguish between the normal class and six distinct CUA categories.
Our contributions is as follows: First, we introduced a dynamic data resampling (DRS) strategy that adjusts sampling weights at both the class and image levels, thereby mitigating the data imbalance caused by disparities in the incidence of uterine anomalies.
Second, we proposed hard sample mining (HSM) strategies to adjust the loss and make the model focus on learning hard cases.  Third, we adopted a self-supervised reconstruction (SSR) method for model pretraining.
This can better explore the limited uterine volumes and enhance the feature modeling ability of CUA-Net.
Last, we developed an online data augmentation (ODA) strategy to refine the prediction during testing, further improving the model’s generalization ability.
Our proposed CUA-Net was validated on a large uterine US dataset, demonstrating strong performance in CUA classification and the potential to transform the traditional clinical workflow (Fig.~\ref{fig:intro}).

\section{Materials and Method}

\subsection{Data source}

Approved by the local IRB (No. KY-KJT-2023-1), we first collected 701 uterus volumes from a medical center between December 2022 and December 2024. All volumes were acquired using a Mindray Resona-9 US system with an intracavitary 3D probe.

Inclusion criteria contain:
(1) Females aged 20-50 years;
(2) Underwent transvaginal 3D US with acceptable image quality;
(3) Diagnosed with normal uterine morphology or CUAs during hysteroscopy or laparoscopy;
(4) Provided informed consent.
Exclusion criteria include: 
(1) Presence of other reproductive system organic diseases (n=12);
(2) Pregnancy/lactation (n=7); 
(3) Use of an intrauterine device (n=21); 
(4) Lack of clinical gold-standard hysteroscopy/laparoscopy supports (n=9).

\begin{figure}[!t]
	\centering
	\includegraphics[width=1.0\linewidth]{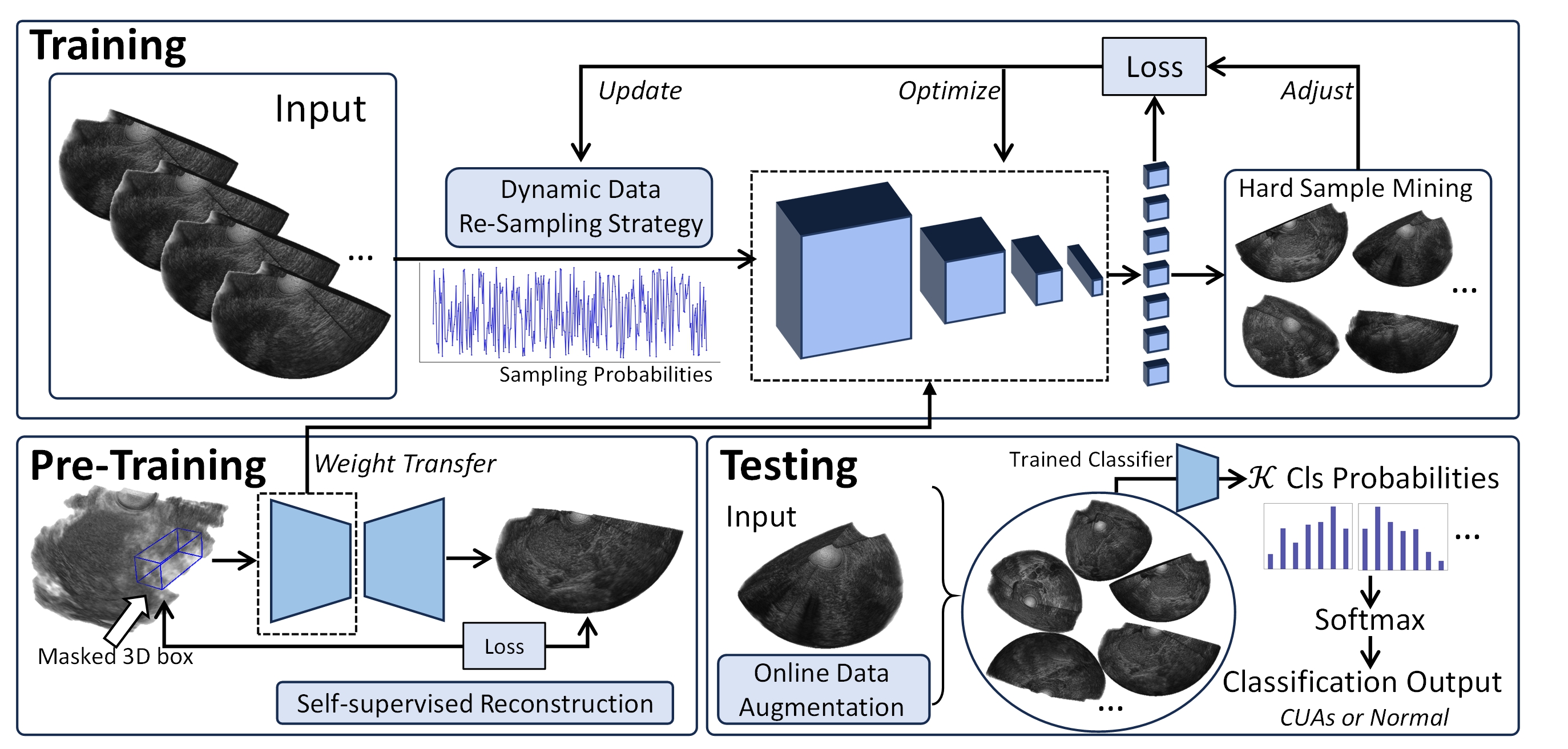}
	\caption{Framework of our proposed CUA-Net.}
	\label{fig:framework}
\end{figure}

After strict data inclusion and exclusion, our dataset for model development included 652 3D US volumes from 444 patients. The dataset comprised 336 normal uterus, 95 septate uterus, 139 arcuate uterus, 49 bicornuate uterus, 17 uterus didelphys, 11 T-shaped uterus, and 5 unicornuate uterus.
The average volume size is 367$\times$189$\times$334, with an isotropic voxel spacing of 0.4$mm$.
Specifically, we randomly split the dataset at patient level into a nearly 6:1:3 ratio for training (388), validation (68), and testing (196).

To enhance the generalizability and robustness of the model, we further incorporated a prospectively collected external test cohort from three hospitals. All data were acquired using GE Voluson E8 US systems between January 2025 and June 2026 and were temporally independent of the model development cohort (before December 2024).
The detailed distribution of uterine types was as follows: 177 normal uterus, 108 septate uterus, 90 arcuate uterus, 36 unicornuate uterus, 31 uterus didelphys, 1 T-shaped uterus, and 5 bicornuate uterus.
Dataset details can be found in Table~\ref{tab:dataset}.

\begin{table}[!t]
  \centering
\caption{Dataset details.}
\label{tab:dataset}
\begin{tabular}{lcccc}
\hline
 & \textbf{Train} & \textbf{Val} & \textbf{Internal Test} & \textbf{External Test}\\
\hline
Normal uterus      & 201 & 34 & 101 & 177\\
Septate uterus     & 56  & 10 & 29  & 108\\
Arcuate uterus     & 83  & 14 & 42  & 90\\
Unicornuate uterus  & 29  & 5  & 15 & 36\\
Uterus didelphys   & 10  & 2  & 5   & 31\\
T-shaped uterus    & 6   & 2  & 3   & 1\\
Bicornuate uterus & 3   & 1  & 1   & 5\\
\hline
\end{tabular}
\end{table}

\subsection{Data pre-processing}
All volumes were proportionally resized to $128 \times 96 \times 128$ and padded with zeros. Online data augmentation during training includes: 
1) random center scaling by $0.9$–$1.1\times$,  
2) random rotation of $0$–$10^\circ$ along the XYZ axes,  
3) random translation of $10$ voxels along the XYZ axes, and  
4) random flipping along all axes.  
Additionally, all data is normalized by dividing by $255$.

\subsection{Model development}

In this study, as shown in Fig.~\ref{fig:framework}, we developed a deep learning-based framework (named CUA-Net) to extract the features from 3D US and directly output the CUA classification results. We used the common 3D ResNet-18 as the model backbone, with a fully connected layer and softmax function, to map the extracted features to the classification probability outputs.
Cross-entropy (CE) was used for training:
\begin{equation}
\mathcal{L}_{\text{CE}} = - \sum_{i=1}^{C} y_i \log(\hat{y}_i),
\end{equation}
where C is the number of classes, $y_i$ is the one-hot GT label and $\hat{y}_i$ is the predicted probability for $i$ class.

\textit{\textbf{Basic Dynamic Data Re-Sampling Strategy.}}
Data resampling is an important strategy to address data imbalance. 
In our task, we first define the class-level base sampling weights inversely proportional to the original sample size of each class (Equ.~\ref{eq:re-sample1}), i.e., $1/n_{y_i}$:
\begin{equation}
    w_{\mathrm{cls}}(y_i) = \frac{1/n_{y_i}}{\sum_{k=1}^{C}(1/n_{y_i})}.
\label{eq:re-sample1}
\end{equation}
We subsequently introduce a dynamic technique to update the image-level probabilities (Equ.~\ref{eq:re-sample2}), driven by the loss function of individual case:
\begin{equation}
    w_{\mathrm{hard}}(x_i) = \frac{\mathrm{loss}(f(x_i), y_i)}{\sum_{j=1}^{N}\mathrm{loss}(f(x_j), y_j)}.
    \label{eq:re-sample2}
\end{equation}

This sampling probability is dynamically adjusted every three iterations, ensuring that samples with higher loss are assigned greater sampling weights.
Finally, the class- and image-level strategies are combined to form the final sampling probability (Equ.~\ref{eq:re-sample3}):
\begin{equation}
    P(x_i) = \frac{w_{\mathrm{cls}}(y_i) \cdot \left(1 + \beta \cdot w_{\mathrm{hard}}(x_i)\right)}{\sum_{j=1}^{N} (w_{\mathrm{cls}}(y_i) \cdot \left(1 + \beta \cdot w_{\mathrm{hard}}(x_i)\right))},
     \label{eq:re-sample3}
\end{equation}
where $\beta$ is a hyperparameter to control the adjustment intensity.

\textit{\textbf{Hard Sample Mining for Loss Adjustment.}}
To fully mine the hard samples and improve the model robustness, we propose a novel hard sample mining strategy.
Simply, sample difficulty can be ranked based on the loss function. However, directly selecting the highest-loss samples may introduce outliers (e.g., noisy data). Therefore, we employ a quantile filtering strategy, retaining high-loss samples while excluding extreme outliers.

\begin{equation}
    \mathcal{L} = \{ \mathcal{L}_{\text{CE}}(x_1), \mathcal{L}_{\text{CE}}(x_2), ..., \mathcal{L}_{\text{CE}}(x_N) \},
\end{equation}

\begin{equation}
    Q_{\tau} = \text{Quantile}(\mathcal{L}, \tau), \quad Q_{\tau'} = \text{Quantile}(\mathcal{L}, \tau'),
\end{equation}

\begin{equation}
    \mathcal{H} = \left\{ x_h \mid Q_{\tau} < \mathcal{L}_{\text{CE}}(x_h) < Q_{\tau'} \right\},
\end{equation}
where $\tau$ and $\tau'$ represents the 80\% and 95\% percentiles, representative. 
$\mathcal{H}$ denotes the hard sample set.
Then, with the update loss weights (Equ.~\ref{eq:hard_weight}),the hard sample driven focal loss can be written as Equ.~\ref{eq:hard_focal}.

\begin{equation}
    w_{\text{hard}}(x_i) = 
    \begin{cases}
        1 + \beta, & x_i \in \mathcal{H} \\
        1, & x_i \notin \mathcal{H}
    \end{cases},
\label{eq:hard_weight}
\end{equation}

\begin{equation}
    \mathcal{L}_{\text{hard-focal}} = - w_{\text{hard}}(x_i) (1 - p_t)^\gamma \log(p_t).
\label{eq:hard_focal}
\end{equation}

Compared to the original focal loss, which amplifies the loss for all low-confidence samples, our improvement better eliminates extreme outliers, reducing noise interference and enhancing learning efficiency.

\textit{\textbf{Self-supervised Reconstruction for Pre-training.}}
Due to the limited CUA data and annotations (classification tags only), we carefully constructed a self-supervised reconstruction method for model pre-training to comprehensively explore the key features in the volume pool.
Inspired by Masked Autoencoder, we first randomly masked a portion of the input data and introduced an encoder-decoder architecture to predict the missing information based only on the visible context.
This can significantly enhance the model's feature representation capability by learning the connection between local and global features, especially in the data scarcity scenario.

Specifically, 3D ResNet-18 was taken as the encoder, while in the decoder part, 3D transposed convolutions were used.
Besides, skip connections were also equipped to link the features of different levels, bringing more spatial information to improve the decoding process.
Different from pixel-level masking, we used the patch-based strategy to reduce the learning difficulty.
We split the volume into non-overlapping patches with size 8$\times$8$\times$8 voxels, producing a total of 3072 patches (128/8$\times$96/8$\times$128/8).
The masking rate was set to 75\%.
We used the common reconstruction loss for training, which calculates errors between predicted ($\hat{x_i}$) and original ($x_i$) images in the masked regions (with $N$ numbers of voxels):
\begin{equation}
\mathcal{L}_{\text{recon}} = \frac{1}{N} \sum_{i=1}^{N} (x_i - \hat{x_i})^2.
\end{equation}
Then, we can inherit the model weights from the above 3D ResNet-18 encoder for the following classification.

\textit{\textbf{Online Data Augmentation Strategy.}}
To enhance model generalization across test cases, we further introduce an online data augmentation strategy.
Specifically, for each test sample $v_i$, we randomly augmented it $\mathcal{K}$ times with the common strategies, including translating, flipping, and rotation, to form a batch input ($\mathcal{K}\times1\times128\times96\times128$).
Then, the classifier will output the probability set with size $\mathcal{K}\times C$, where $C=7$ indicates the number of CUA types.
Through the \textit{average} operation at the class channel and after a $softmax$, an optimized probability $p_o$ is obtained.
Then, the predicted tag for the current testing uterus volume $v_i$ is defined as $c^*(v_i)=argmax(p_o)$.

\subsection{Experimental setting}
All experiments were implemented using Python (version 3.8.0) and PyTorch (version 2.0.0) on an NVIDIA A40 GPU with 48GB of memory. 
The self-supervised reconstruction was trained using the Adam optimizer with a learning rate of $1 \times 10^{-4}$ for 30 epochs.  
For classification, we also used the Adam optimizer with an initial learning rate of $1 \times 10^{-3}$.  
The learning rate follows a linear decay schedule, decreasing by $10\%$ every 40 epochs.  
The total number of epochs is set to 100.  
Models achieving the best performance on the validation set were selected for final evaluation.

\subsection{Statistical analysis}
We evaluated the model using different metrics, including Accuracy, Precision, Recall, F1-score, micro-AUC, macro-AUC, etc.
95\% CIs were evaluated by bootstrapping with 1,000 resamples.
For statistical analysis, Accuracy was compared using the paired McNemar test, while differences in other metrics were assessed using two-sided paired permutation tests with 1,000 permutations.
Moreover, for comparison between CUA-Net and sonographers, we used Bonferroni correction to update the $p$-values and reduce false positive rate. The adjusted p-value was calculated as:
\begin{equation}
    p_{\mathrm{cor}}
    =\min\left(N_{\mathrm{sono}}*N_{\mathrm{metrics}}*p_{\mathrm{raw}},1\right),
\end{equation}
where $N_{\mathrm{sono}}=2$ and $N_{\mathrm{metrics}}=4$ denote the numbers of sonographers and evaluated performance metrics within each comparison group, respectively. Here, $p_{\mathrm{cor}}$ and $p_{\mathrm{raw}}$ denote the corrected and raw p-values.
$p_{\mathrm{cor}}<$0.05 indicates a statistically significant performance difference.

\section{Results}
To validate the effectiveness of the proposed method, we compared CUA-Net with 2D/3D ResNet-18 and several mainstream advanced video/3D models.
As shown in Table~\ref{tab:comparison_two_rows}, CUA-Net achieved the best performance across all evaluation metrics, e.g., accuracy of 93.88\% (95\% CI: 89.80–97.45), and F1-score of 88.09\% (95\% CI: 69.73–92.79), etc.
Compared with the second-best model 3D nnMamba, it improved accuracy by 9.70\% and macro-AUC by 0.0026.
Compared with 2D ResNet-18, 3D ResNet-18 achieved significantly better accuracy (75.00\% vs. 42.86\%, $p<0.05$) and macro-AUC (0.9566 vs. 0.9108, $p<0.05$), verifying the advantages of 3D modeling.
Accordingly, considering both performance and model complexity, we adopted the relatively lightweight 3D ResNet-18, with 33.2M parameters, as the backbone for all subsequent experiments.

\begin{table}[!t]
\centering
\caption{Comparison of different models on the CUA classification task.
$^{*}$ indicates statistical significance compared with all other methods ($p<0.05$)}
\label{tab:comparison_two_rows}
\resizebox{\textwidth}{!}{%
\begin{tabular}{lcccccc}
\toprule
\textbf{Methods} & \textbf{Accuracy (\%)} & \textbf{Precision (\%)} & \textbf{Recall (\%)} & \textbf{F1 (\%)} & \textbf{micro-AUC} & \textbf{macro-AUC} \\
\midrule

2D ResNet-18
& \makecell{42.86 \\ (36.20--50.00)}
& \makecell{42.85 \\ (26.24--47.17)}
& \makecell{30.52 \\ (22.35--37.95)}
& \makecell{28.98 \\ (17.90--35.95)}
& \makecell{0.8280 \\ (0.7872--0.8620)}
& \makecell{0.9108 \\ (0.8489--0.9532)} \\

I3D
& \makecell{69.90 \\ (63.51--76.02)}
& \makecell{57.23 \\ (51.24--62.72)}
& \makecell{46.77 \\ (41.16--50.81)}
& \makecell{44.15 \\ (35.89--50.61)}
& \makecell{0.9519 \\ (0.9332--0.9677)}
& \makecell{0.9820 \\ (0.9730--0.9905)} \\

Video Swin transformer
& \makecell{75.00 \\ (69.11--80.88)}
& \makecell{52.86 \\ (50.16--56.01)}
& \makecell{56.42 \\ (51.52--61.15)}
& \makecell{47.83 \\ (41.55--53.61)}
& \makecell{0.9660 \\ (0.9525--0.9769)}
& \makecell{0.9965 \\ (0.9930--0.9991)} \\

3D ResNet-18 (Baseline)
& \makecell{72.45 \\ (65.82--78.33)}
& \makecell{55.17 \\ (37.80--60.14)}
& \makecell{49.11 \\ (44.47--52.40)}
& \makecell{40.95 \\ (31.95--47.12)}
& \makecell{0.9342 \\ (0.9076--0.9599)}
& \makecell{0.9566 \\ (0.9296--0.9784)} \\

3D MedicalNet
& \makecell{81.12 \\ (76.02--86.22)}
& \makecell{52.83 \\ (46.82--59.12)}
& \makecell{69.05 \\ (54.73--70.73)}
& \makecell{57.39 \\ (50.53--63.46)}
& \makecell{0.9172 \\ (0.8796--0.9450)}
& \makecell{0.9946 \\ (0.9906--0.9975)} \\

3D DenseNet
& \makecell{82.65 \\ (76.77--87.76)}
& \makecell{69.80 \\ (52.06--74.99)}
& \makecell{75.24 \\ (58.18--77.74)}
& \makecell{68.06 \\ (48.82--73.08)}
& \makecell{0.9841 \\ (0.9759--0.9910)}
& \makecell{0.9996 \\ (0.9987--0.9999)} \\

3D ViT
& \makecell{78.06 \\ (71.93--83.16)}
& \makecell{82.56 \\ (51.91--86.95)}
& \makecell{76.65 \\ (59.90--79.19)}
& \makecell{68.14 \\ (48.20--73.47)}
& \makecell{0.9796 \\ (0.9674--0.9877)}
& \makecell{0.9988 \\ (0.9972--0.9998)} \\

3D nnMamba
& \makecell{84.18 \\ (78.30--89.04)}
& \makecell{85.31 \\ (69.39--87.33)}
& \makecell{80.81 \\ (53.95--86.85)}
& \makecell{75.59 \\ (56.34--81.05)}
& \makecell{0.9869 \\ (0.9800--0.9925)}
& \makecell{0.9971 \\ (0.9948--0.9987)} \\

CUA-Net
& \makecell{\textbf{93.88$^{*}$} \\ \textbf{(89.80--97.45)}}
& \makecell{\textbf{87.01$^{*}$} \\ \textbf{(69.51--91.30)}}
& \makecell{\textbf{95.92$^{*}$} \\ \textbf{(79.85--97.68)}}
& \makecell{\textbf{88.09$^{*}$} \\ \textbf{(69.73--92.79)}}
& \makecell{\textbf{0.9982$^{*}$} \\ \textbf{(0.9963--0.9995)}}
& \makecell{\textbf{0.9997$^{*}$} \\ \textbf{(0.9990--1.0000)}} \\

\bottomrule
\end{tabular}%
}
\end{table}

As shown in Table~\ref{tab:method_comparison}, the naive 3D ResNet-18 achieves limited performance (Accuracy 72.45\%, F1: 40.95\%), falling short of clinical intelligent CUA classification requirements.
Adding the DRS and HSM can improve the model performance, validating their efficacy in handling the data imbalance and hard sample learning issues.
However, because hard samples are selected globally between the 80th and 95th loss percentiles, optimization may become biased toward difficult cases from certain classes, temporarily affecting class-wise probability ranking. 
Moreover, some samples within the highest-loss group may represent informative rare/atypical cases rather than true noise. 
Excluding them may weaken class-specific discrimination and consequently reduce macro-AUC, which weights all classes equally, despite improvements in accuracy and micro-AUC.
This effect was subsequently alleviated by SSR and ODA, which enhanced feature learning and improved prediction stability.
Specifically, adding SSR boosts all metrics, including the previously-dropped precision, macro-AUC, etc.
This proves that the self-supervised technique assists the model in fully utilizing limited data and capturing classification-relevant relationships between local patches and global volumes.
Incorporating ODA also enhances CUA analysis, allowing the model to observe and analyze volumes from different perspectives through online learning, ultimately refining classification decisions via self-adjustment.
Finally, CUA-Net achieves an accuracy of 93.88\% and an F1-score of 88.09\%, further validating the overall contributions of the proposed strategies.
\begin{table}[!h]
\centering
\caption{Ablation study.}
\label{tab:method_comparison}
\resizebox{\textwidth}{!}{%
\begin{tabular}{lcccccc}
\toprule
\textbf{Methods} & \textbf{Accuracy (\%)} & \textbf{Precision (\%)} & \textbf{Recall (\%)} & \textbf{F1 (\%)} & \textbf{micro-AUC}& \textbf{macro-AUC}\\\midrule 

\textit{Baseline}
& \makecell{72.45 \\ (65.82-78.33)}
& \makecell{55.17 \\ (37.80-60.14)}
& \makecell{49.11 \\ (44.47-52.40)}
& \makecell{40.95 \\ (31.95-47.12)}
& \makecell{0.9342 \\ (0.9076-0.9599)}
& \makecell{0.9566 \\ (0.9296-0.9784)} \\ 

\textit{Baseline+DRS}
& \makecell{80.10 \\ (74.49-84.96)}
& \makecell{74.97 \\ (57.31-78.83)}
& \makecell{61.20 \\ (53.60-71.25)}
& \makecell{62.04 \\ (50.17-70.97)}
& \makecell{0.9814 \\ (0.9719-0.9874)}
& \makecell{0.9759 \\ (0.9672-0.9825)} \\ 

\textit{Baseline+DRS+HSM}
& \makecell{86.22 \\ (81.86-90.31)}
& \makecell{59.52 \\ (57.78-61.90)}
& \makecell{62.35 \\ (58.21-65.73)}
& \makecell{56.21 \\ (51.90-60.10)}
& \makecell{0.9871 \\ (0.9782-0.9922)}
& \makecell{0.9110 \\ (0.8983-0.9212)} \\ 

\textit{Baseline+DRS+HSM+SSR}
& \makecell{90.31 \\ (86.47-94.39)}
& \makecell{75.78 \\ (59.91-80.67)}
& \makecell{78.96 \\ (64.06-82.32)}
& \makecell{76.34 \\ (60.99-80.56)}
& \makecell{0.9898 \\ (0.9821-0.9961)}
& \makecell{0.9967 \\ (0.9932-0.9995)} \\ 

\textit{Baseline+DRS+HSM+SSR+ODA (CUA-Net)}
& \makecell{93.88 \\ (89.80-97.45)}
& \makecell{87.01 \\ (69.51-91.30)}
& \makecell{95.92 \\ (79.85-97.68)}
& \makecell{88.09 \\ (69.73-92.79)}
& \makecell{0.9982 \\ (0.9963-0.9995)}
& \makecell{0.9997 \\ (0.9990-1.0000)} \\
\bottomrule
\end{tabular}}
\end{table}

\begin{figure}[!t]
	\centering
\includegraphics[width=1.0\linewidth]{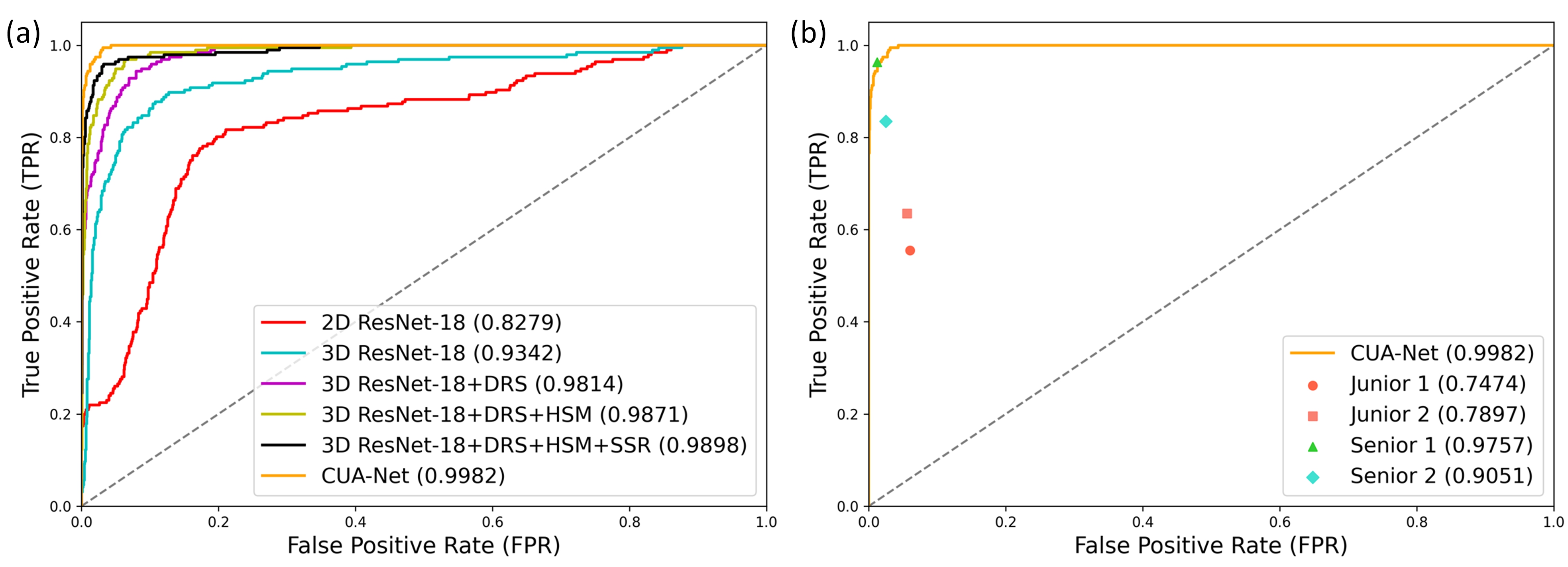}
	\caption{ROC curves with AUC values for (a) different methods and (b) junior and senior sonographers. The AUC values for sonographers are approximated by connecting several points, i.e., (0,0), (FPR, TPR), and (1,1), following~\cite{zhou2021ensembled}.}
	\label{fig:roc}
\end{figure}

\begin{figure}[!t]
	\centering
	\includegraphics[width=1.0\linewidth]{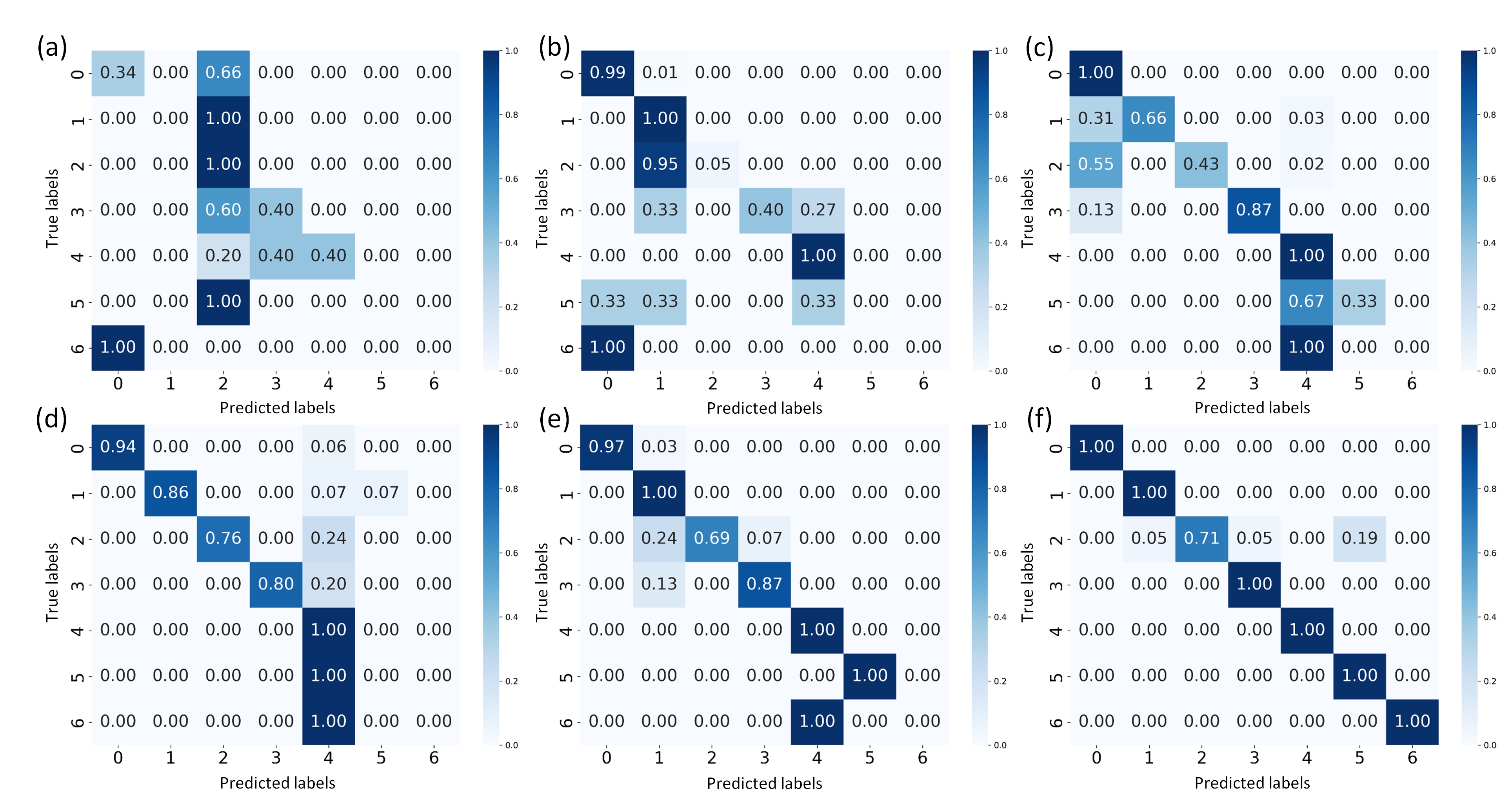}
	\caption{Confusion matrix of different methods, including (a) 2D ResNet-18, (b) 3D ResNet-18, 3D ResNet-18 with (c) DRS, (d) DRS+HSM, (e) DRS+HSM+SSR, and (f) DRS+HSM+SSR+ODA. 0-6 correspond to normal uterus, septate uterus, arcuate uterus, uterus didelphys, unicornuate uterus, T-shaped uterus, and bicornuate uterus, respectively.}
	\label{fig:cm}
\end{figure}

As shown in Fig.~\ref{fig:roc} (a), the ROC curves with AUC values (same as in Table~\ref{tab:dataset}) of different methods further prove the strong overall performance of our method under various thresholds.
In Fig.~\ref{fig:cm}, we visualize the confusion matrix of different methods to show the class-level analysis.
It further validates the superiority of our CUA-Net (see Fig.~\ref{fig:cm} (f)) on all classes, compared to other methods.
We noticed that the Arcuate uterus (True label=2) often exhibits poorer performance than others.
This may be due to its similar characteristics to other uterine types, which can confuse network learning.

\begin{figure}[!t]
	\centering
	\includegraphics[width=1.0\linewidth]{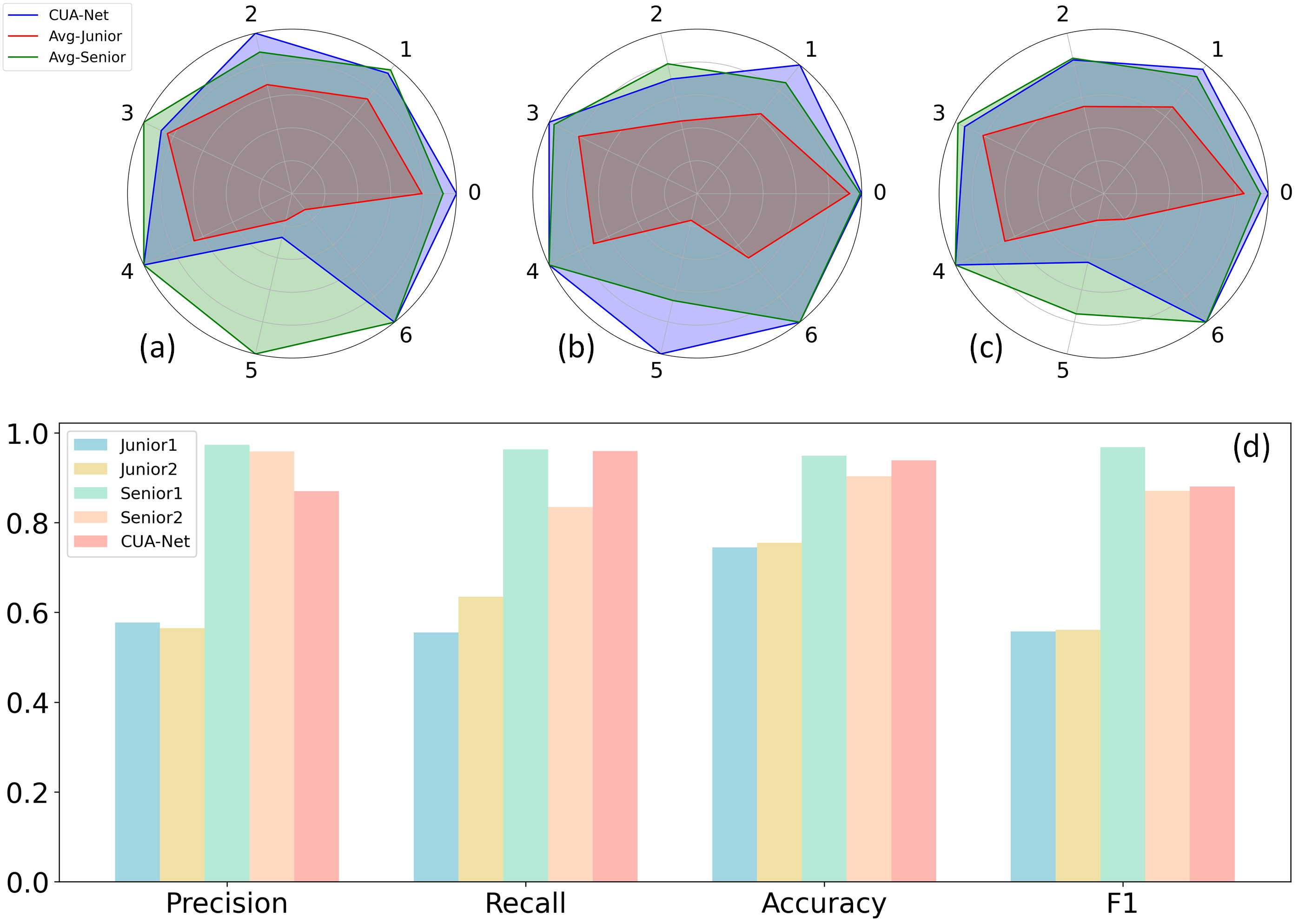}
	\caption{Performance comparison between CUA-Net and junior/senior sonographers.
    Radar charts illustrate (a) precision, (b) recall, and (c) F1-score across different categories (0-6: normal uterus, septate uterus, arcuate uterus, uterus didelphys, unicornuate uterus, T-shaped uterus and bicornuate uterus) for CUA-Net (blue), average juniors (red), and average seniors (green), respectively. 
    (d) Bar chart presents the overall precision, recall, accuracy, and F1-score of juniors (Junior1/2), seniors (Senior1/2), and CUA-Net.}
	\label{fig:rader}
\end{figure}

We also compared the CUA-Net with two sonographer groups: one involving two junior sonographers (\textless5 years of experience) and one two senior sonographers (\textgreater15 years of experience).
Within each group, a total of eight hypothesis tests were conducted, comprising two sonographer comparisons across four performance metrics (i.e., $p_{cor}=p_{raw}/8$).
As shown in Table~\ref{tab:sonographer_comparison} and Fig.~\ref{fig:rader}, CUA-Net achieved higher accuracy, precision, recall, and F1-score than both junior sonographers. After Bonferroni correction, all differences remained statistically significant (corrected $p$-values$<0.05$), indicating that CUA-Net consistently outperformed both junior sonographers across all evaluated metrics.
For the senior group, no statistically significant differences were observed between CUA-Net and the two senior sonographers in most comparisons after Bonferroni correction (i.e., 7/8 metrics with corrected $p$-values$>0.05$, only Precision for Senior 1 $vs.$ CUA-Net with corrected $p$-values=0.04). 
Overall, these results suggest that CUA-Net outperformed junior sonographers, and achieved competitive performance relative to the senior sonographers across most evaluated metrics.

\begin{table}[!t]
\centering
\caption{Performance comparison between CUA-Net and sonographers.}
\label{tab:sonographer_comparison}
\resizebox{!}{0.2\textwidth}{
\begin{tabular}{lcccc}
\toprule
Methods & Accuracy (\%) & Precision (\%) & Recall (\%) & F1 (\%) \\
\midrule

\multirow{2}{*}{Junior 1}
& 74.49 & 57.73 & 55.50 & 55.76 \\
& (67.86--81.12)
& (47.12--71.61)
& (43.88--69.64)
& (44.34--67.79) \\

\multirow{2}{*}{Junior 2}
& 75.51 & 56.48 & 63.51 & 56.10 \\
& (68.88--81.63)
& (43.78--68.98)
& (42.34--71.43)
& (42.39--67.76) \\

\multirow{2}{*}{Senior 1}
& 94.90 & 97.36 & 96.34 & 96.82 \\
& (91.84--97.96)
& (94.83--99.01)
& (93.35--98.42)
& (94.02--98.65) \\

\multirow{2}{*}{Senior 2}
& 90.31 & 95.85 & 83.52 & 87.13 \\
& (86.22--94.39)
& (76.87--97.45)
& (72.79--93.58)
& (74.57--94.81) \\

\multirow{2}{*}{CUA-Net}
& 93.88 & 87.01 & 95.92 & 88.09 \\
& (89.80--97.45)
& (69.51--91.30)
& (79.85--97.68)
& (69.73--92.79) \\

\bottomrule
\end{tabular}}
\end{table}

\begin{figure}[!t]
	\centering
	\includegraphics[width=1.0\linewidth]{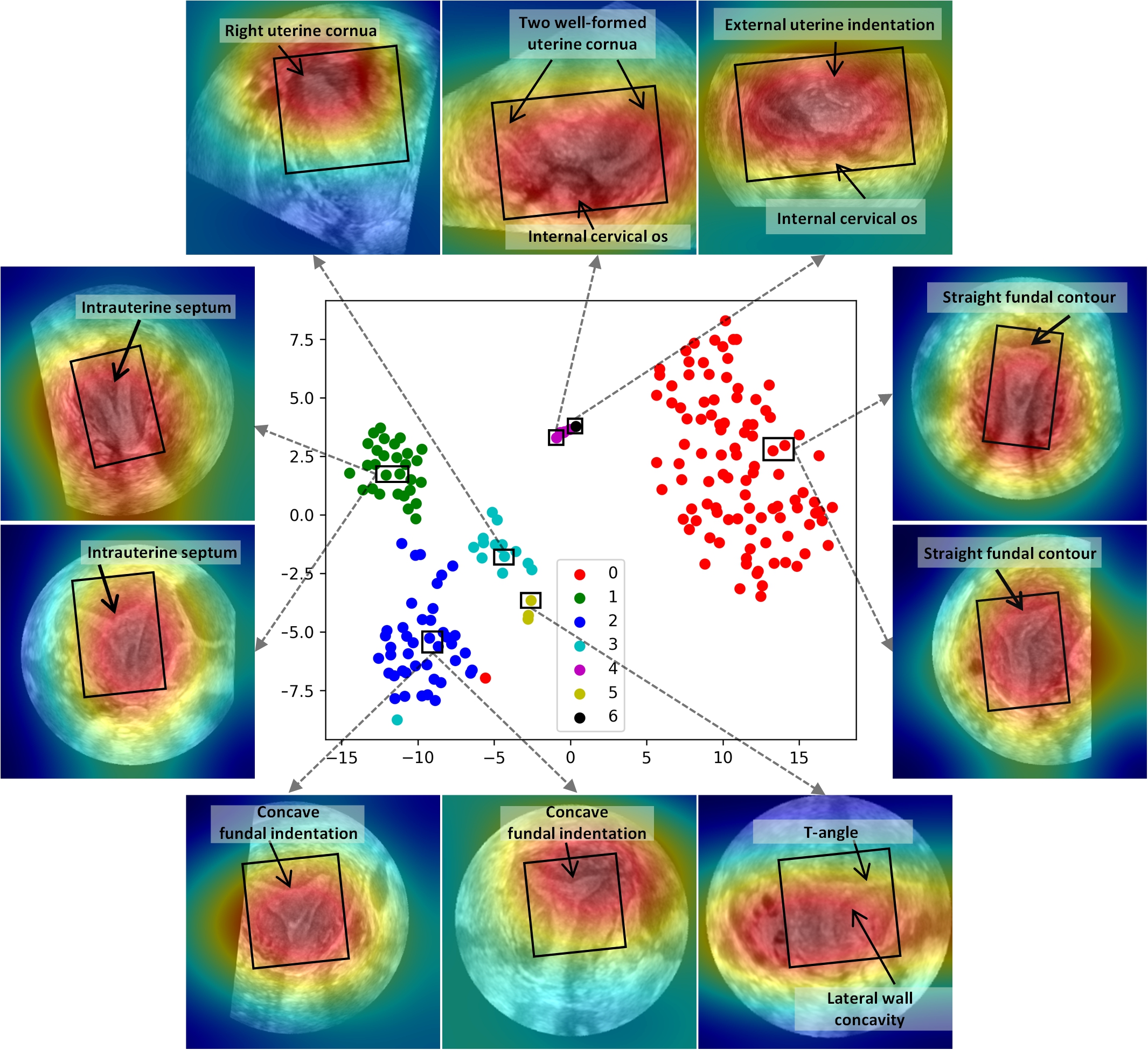}
	\caption{t-SNE visualization with CAM results indicating the attention regions by our CUA-Net. The black bounding boxes show the key regions provided by senior sonographers, and the black arrows point out the detailed structures. The categories 0 to 6 correspond to normal uterus, septate uterus, arcuate uterus, uterus didelphys, unicornuate uterus, T-shaped uterus and bicornuate uterus, respectively.}
	\label{fig:cam}
\end{figure}

Fig.~\ref{fig:cam} presents qualitative visualization results based on t-SNE and Class Activation Mapping (CAM).
The t-SNE visualization reveals that our method achieves effective feature modeling, successfully learning discriminative representations for different categories.
In addition, we applied the Grad-CAM technique on the testing set to examine the network’s attention regions across different cases. 
Specifically, Grad-CAM was first used to generate 3D heatmaps, and then, based on additional annotated coronal plane parameters, the corresponding 2D slices and their projected Grad-CAM responses were extracted.
See the 10 pseudo-color maps in Fig.~\ref{fig:cam}, the redder the color, the higher the level of attention from the model, and the black boxes highlight the regions that senior sonographers typically focus on.
We also report quantitative attention-alignment analyses using the Dice similarity coefficient (Dice) and intersection over union (IoU), as shown in Table~\ref{tab:attention_alignment}.
It should be noted that CUA-Net is a classification model trained using only category labels, without any segmentation/detection annotations. Hence, the attention maps were generated in a weakly supervised manner. 
The quantitative Dice and IoU results show a moderate degree of overlap between the regions highlighted by CUA-Net and the diagnostic regions of interest identified by sonographers across different uterine categories. 
These findings indicate that CUA-Net can focus on clinically relevant anatomical regions to make decision, further supporting its effectiveness and interpretability.

\begin{table}[!t]
\centering
\caption{Quantitative evaluation of attention alignment across different uterine categories. Values are reported as mean $\pm$ standard deviation.}
\label{tab:attention_alignment}
\begin{tabular}{lcc}
\toprule
\textbf{Category} & \textbf{Dice} & \textbf{IoU} \\
\midrule
Normal uterus
& $0.7525 \pm 0.0854$
& $0.6879 \pm 0.1326$ \\

Septate uterus
& $0.7213 \pm 0.0976$
& $0.6358 \pm 0.1412$ \\

Arcuate uterus
& $0.6847 \pm 0.1128$
& $0.5875 \pm 0.1534$ \\

Unicornuate uterus
& $0.7389 \pm 0.0931$
& $0.6542 \pm 0.1375$ \\

Uterus didelphys
& $0.7214 \pm 0.1013$
& $0.6210 \pm 0.1387$ \\

T-shaped uterus
& $0.7024 \pm 0.1059$
& $0.6087 \pm 0.1486$ \\

Bicornuate uterus
& $0.6598 \pm 0.1286$
& $0.5845 \pm 0.1511$ \\

\bottomrule
\end{tabular}
\end{table}

\begin{table}[!t]
\centering
\caption{
Performance comparison between CUA-Net and competing methods on the independent external test set.
Best performances are highlighted in bold, and $^{*}$ indicates statistical significance compared with all other methods ($p<0.05$).
Red and blue values indicate increases and decreases, respectively, relative to the internal test set in Table~\ref{tab:method_comparison}.
}
\label{tab:external_comparison}
\resizebox{\textwidth}{!}{
\begin{tabular}{lcccccc}
\toprule
\textbf{Methods}
& \textbf{Accuracy (\%)}
& \textbf{Precision (\%)}
& \textbf{Recall (\%)}
& \textbf{F1-score (\%)}
& \textbf{Micro-AUC}
& \textbf{Macro-AUC} \\
\midrule

2D ResNet-18
& \makecell{39.96\textcolor{blue}{\scriptsize$_{2.90\downarrow}$}\\(34.93--44.42)}
& \makecell{12.84\textcolor{blue}{\scriptsize$_{30.01\downarrow}$}\\(5.39--20.37)}
& \makecell{15.21\textcolor{blue}{\scriptsize$_{15.31\downarrow}$}\\(14.29--16.75)}
& \makecell{9.78\textcolor{blue}{\scriptsize$_{19.20\downarrow}$}\\(7.83--12.40)}
& \makecell{0.7153\textcolor{blue}{\scriptsize$_{0.1127\downarrow}$}\\(0.6889--0.7418)}
& \makecell{0.6219\textcolor{blue}{\scriptsize$_{0.2889\downarrow}$}\\(0.5703--0.6614)} \\

I3D
& \makecell{55.58\textcolor{blue}{\scriptsize$_{14.32\downarrow}$}\\(51.56--59.60)}
& \makecell{51.27\textcolor{blue}{\scriptsize$_{5.96\downarrow}$}\\(47.81--54.52)}
& \makecell{32.37\textcolor{blue}{\scriptsize$_{14.40\downarrow}$}\\(28.98--35.50)}
& \makecell{34.98\textcolor{blue}{\scriptsize$_{9.17\downarrow}$}\\(30.44--38.30)}
& \makecell{0.8492\textcolor{blue}{\scriptsize$_{0.1027\downarrow}$}\\(0.8206--0.8724)}
& \makecell{0.7325\textcolor{blue}{\scriptsize$_{0.2495\downarrow}$}\\(0.6654--0.7868)} \\

Video Swin Transformer
& \makecell{55.34\textcolor{blue}{\scriptsize$_{19.66\downarrow}$}\\(50.22--59.38)}
& \makecell{55.58\textcolor{red}{\scriptsize$_{2.72\uparrow}$}\\(53.87--56.93)}
& \makecell{41.37\textcolor{blue}{\scriptsize$_{15.05\downarrow}$}\\(38.56--44.37)}
& \makecell{37.14\textcolor{blue}{\scriptsize$_{10.69\downarrow}$}\\(33.34--40.15)}
& \makecell{0.8822\textcolor{blue}{\scriptsize$_{0.0838\downarrow}$}\\(0.8641--0.9010)}
& \makecell{0.8650\textcolor{blue}{\scriptsize$_{0.1315\downarrow}$}\\(0.8142--0.9233)} \\

3D ResNet-18 (Baseline)
& \makecell{68.53\textcolor{blue}{\scriptsize$_{3.92\downarrow}$}\\(64.39--72.77)}
& \makecell{51.09\textcolor{blue}{\scriptsize$_{4.08\downarrow}$}\\(47.88--54.18)}
& \makecell{41.12\textcolor{blue}{\scriptsize$_{7.99\downarrow}$}\\(37.58--44.73)}
& \makecell{42.44\textcolor{red}{\scriptsize$_{1.49\uparrow}$}\\(37.82--45.92)}
& \makecell{0.9322\textcolor{blue}{\scriptsize$_{0.0020\downarrow}$}\\(0.9150--0.9456)}
& \makecell{0.8806\textcolor{blue}{\scriptsize$_{0.0760\downarrow}$}\\(0.8584--0.9008)} \\

3D MedicalNet
& \makecell{59.15\textcolor{blue}{\scriptsize$_{21.97\downarrow}$}\\(54.69--63.39)}
& \makecell{60.97\textcolor{red}{\scriptsize$_{8.14\uparrow}$}\\(58.71--63.30)}
& \makecell{36.72\textcolor{blue}{\scriptsize$_{32.33\downarrow}$}\\(33.50--40.26)}
& \makecell{37.70\textcolor{blue}{\scriptsize$_{19.69\downarrow}$}\\(33.64--41.61)}
& \makecell{0.9048\textcolor{blue}{\scriptsize$_{0.0124\downarrow}$}\\(0.8848--0.9205)}
& \makecell{0.8369\textcolor{blue}{\scriptsize$_{0.1577\downarrow}$}\\(0.8059--0.8644)} \\

3D DenseNet
& \makecell{77.68\textcolor{blue}{\scriptsize$_{4.97\downarrow}$}\\(73.88--81.59)}
& \makecell{60.82\textcolor{blue}{\scriptsize$_{8.98\downarrow}$}\\(51.58--69.13)}
& \makecell{61.13\textcolor{blue}{\scriptsize$_{14.11\downarrow}$}\\(56.60--67.20)}
& \makecell{56.24\textcolor{blue}{\scriptsize$_{11.82\downarrow}$}\\(49.46--62.40)}
& \makecell{0.9683\textcolor{blue}{\scriptsize$_{0.0158\downarrow}$}\\(0.9581--0.9768)}
& \makecell{0.9489\textcolor{blue}{\scriptsize$_{0.0507\downarrow}$}\\(0.9017--0.9812)} \\

3D ViT
& \makecell{75.00\textcolor{blue}{\scriptsize$_{3.06\downarrow}$}\\(70.98--78.57)}
& \makecell{74.33\textcolor{blue}{\scriptsize$_{8.23\downarrow}$}\\(63.75--79.11)}
& \makecell{60.45\textcolor{blue}{\scriptsize$_{16.20\downarrow}$}\\(51.38--67.73)}
& \makecell{63.47\textcolor{blue}{\scriptsize$_{4.67\downarrow}$}\\(53.01--69.11)}
& \makecell{0.9675\textcolor{blue}{\scriptsize$_{0.0121\downarrow}$}\\(0.9569--0.9744)}
& \makecell{0.9262\textcolor{blue}{\scriptsize$_{0.0726\downarrow}$}\\(0.9184--0.9334)} \\

3D nnMamba
& \makecell{76.11\textcolor{blue}{\scriptsize$_{8.07\downarrow}$}\\(71.65--79.91)}
& \makecell{79.65\textcolor{blue}{\scriptsize$_{5.66\downarrow}$}\\(78.52--80.73)}
& \makecell{70.03\textcolor{blue}{\scriptsize$_{10.78\downarrow}$}\\(67.43--72.02)}
& \makecell{71.02\textcolor{blue}{\scriptsize$_{4.57\downarrow}$}\\(68.29--73.19)}
& \makecell{0.9649\textcolor{blue}{\scriptsize$_{0.0220\downarrow}$}\\(0.9540--0.9735)}
& \makecell{0.9437\textcolor{blue}{\scriptsize$_{0.0534\downarrow}$}\\(0.9368--0.9507)} \\

\textbf{CUA-Net}
& \makecell{\textbf{91.52}$^{*}$\textcolor{blue}{\scriptsize$_{2.36\downarrow}$}\\\textbf{(89.06--94.20)}}
& \makecell{\textbf{83.27}$^{*}$\textcolor{blue}{\scriptsize$_{3.74\downarrow}$}\\\textbf{(69.18--93.77)}}
& \makecell{\textbf{88.63}$^{*}$\textcolor{blue}{\scriptsize$_{7.29\downarrow}$}\\\textbf{(68.73--95.14)}}
& \makecell{\textbf{81.49}$^{*}$\textcolor{blue}{\scriptsize$_{6.60\downarrow}$}\\\textbf{(67.71--91.26)}}
& \makecell{\textbf{0.9945}$^{*}$\textcolor{blue}{\scriptsize$_{0.0037\downarrow}$}\\\textbf{(0.9917--0.9967)}}
& \makecell{\textbf{0.9990}$^{*}$\textcolor{blue}{\scriptsize$_{0.0007\downarrow}$}\\\textbf{(0.9976--0.9998)}} \\

\bottomrule
\end{tabular}
}
\end{table}

\begin{figure}[!h]
	\centering
	\includegraphics[width=\linewidth]{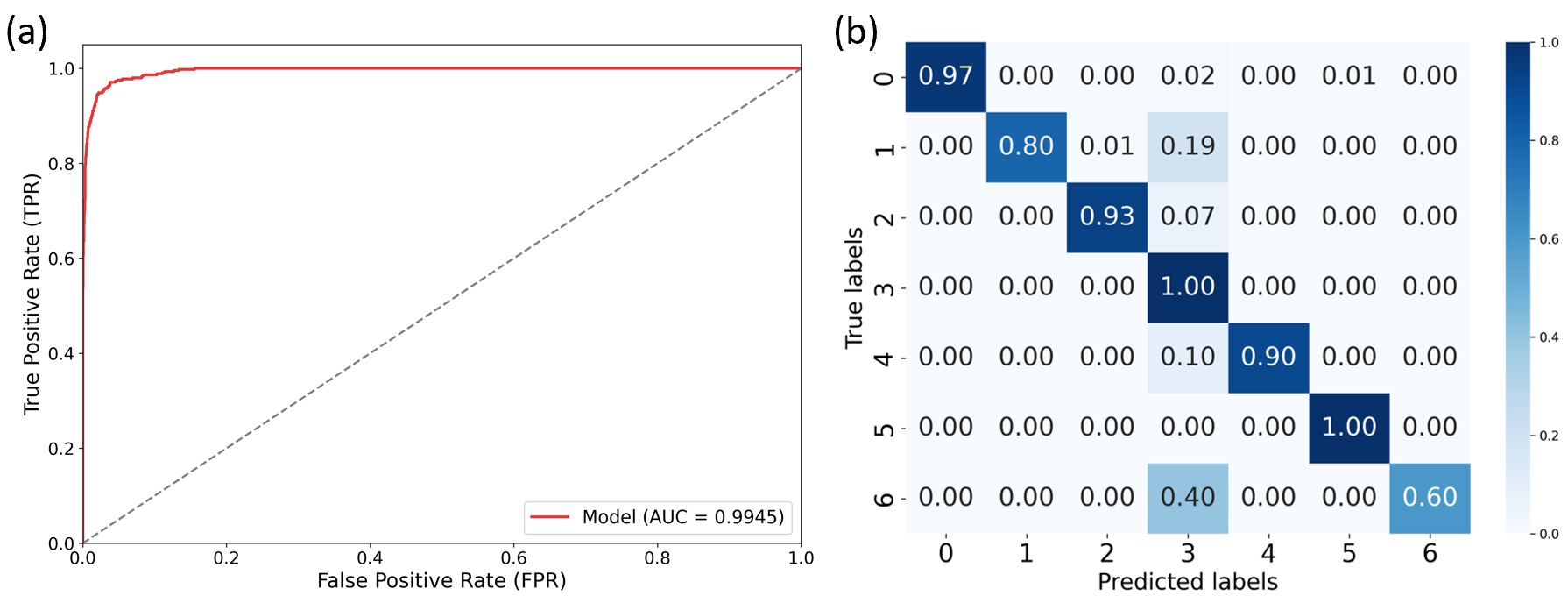}
	\caption{(a) ROC curves with AUC values and (b) the confusion matrix of CUA-Net on the external test cohort. 0-6 denotes: normal uterus, septate uterus, arcuate uterus, unicornuate uterus, uterus didelphys, T-shaped uterus,  bicornuate uterus, respectively.}
	\label{fig:results_etest}
\end{figure}

To further verify the generalizability of the proposed CUA-Net, we evaluated its performance on an external test cohort.
As shown in Table~\ref{tab:external_comparison}, we compared CUA-Net with eight competing methods.
It achieved the best performance across all evaluation metrics, with an accuracy of 91.52\%, an F1-score of 81.49\%, etc (detailed performance refers to Fig.~\ref{fig:results_etest}). These improvements were statistically significant compared with all competing methods ($p<0.05$).
Compared with the average declines of 9.86, 6.51, 15.77, and 9.79 percentage points in accuracy, precision, recall, and F1-score and 0.0454 and 0.1350 in micro- and macro-AUC across the eight competing methods (from internal to external testing), CUA-Net showed markedly smaller drops of 2.36, 3.74, 7.29, and 6.60 percentage points and 0.0037 and 0.0007, respectively.
Moreover, the class-wise confusion matrix demonstrates reliable generalization across common CUA categories, while also showing preliminary potential for recognizing rare anomalies such as unicornuate, bicornuate, and T-shaped uterus (Fig.~\ref{fig:results_etest}).
These results further support CUA-Net's acceptable robustness and generalization to independent external data.

\section{Discussion}
CUA is among the most common gynecological disorders, with a relatively high prevalence rate.
It is closely associated with female infertility and may also lead to reproductive complications such as recurrent miscarriage, intrauterine growth restriction, preterm birth, and retained placenta.
Therefore, accurate CUA classification is crucial for selecting appropriate surgical approaches and related treatments.
US is a primary tool for diagnosing CUAs; however, conventional 2D US cannot capture the coronal plane, whereas 3D US visualizes the coronal plane, providing a comprehensive view of uterine structures to drive accurate diagnosis.
Specifically, in traditional clinical workflow, sonographers must first locate and reconstruct the 2D standard coronal plane that meets diagnostic criteria within the high-dimensional 3D space, and then perform measurements on this plane to obtain an accurate CUA diagnosis.
This process is highly dependent on the clinician’s experience, time-consuming, and negatively impacts diagnostic efficiency.

With the widespread adoption of AI, its integration into medical ultrasound image analysis has continued to advance.
Several deep learning-based methods have been introduced to achieve intelligent analysis in 3D uterine US. Most of them focus solely on detecting standard planes in 3D space~\cite{yang2021searching,zou2022agent,huang2023localizing}, neglecting the vital CUA classification task. Two recent works explored classifying (1) normal/abnormal cases~\cite{dou2025standard} and (2) normal/CUA samples~\cite{huang2025uncertainty} based on deep learning techniques.
However, they still highly rely on the plane localization process, which may degrade the diagnosis performance due to the potential error propagation, e.g., a wrongly located plane will very likely lead to misclassification of CUA.
Hence, designing a deep learning-based model that directly processes 3D uterine data and outputs CUA predictions in one step could improve overall accuracy by avoiding error accumulation. This could also reshape the clinical workflow and greatly enhance diagnostic effectiveness.

In this paper, we built the CUA-Net to achieve the above goal.
Our proposed CUA-Net builds upon the baseline, i.e., 3D ResNet-18, and achieves significant performance improvements through the stepwise integration of data re-sampling (DRS),  hard sample mining (HSM), self-supervised reconstruction (SSR), and online data augmentation (ODA) techniques.
Due to their low incidence in the real world, complex CUAs such as T-shaped uterus and bicornuate uterus are underrepresented in our collected dataset.
Our proposed strategies can alleviate the class imbalance issue, learn from the difficult cases, extract features from the limited volumes, and correct the wrong predictions during testing, to comprehensively improve the model performance on CUA classification.
In conclusion, our CUA-Net achieved the best performance among different competitors on all metrics, including Accuracy (93.88\%), Precision (87.01\%), Recall (95.92\%), F1 (88.09\%), Micro-AUC (0.9982), and Macro-AUC (0.9997).

To assess clinical applicability, we compared CUA-Net with both junior and senior sonographers.
CUA-Net outperformed junior sonographers across all metrics, demonstrating its potential to assist them in CUA classification. 
CUA-Net also demonstrated performance comparable to that of senior sonographers, with no statistically significant differences observed in most metrics.
Besides, CUA-Net only required about 0.1s for testing, significantly faster than juniors (61.8s) and seniors (40.6s).
This demonstrates the model’s efficiency and effectiveness, enabling real-time clinical classification.

Regarding the limitations, CUA-Net may still misclassify certain CUA categories. For example, it showed relatively poor performance in identifying arcuate uterus (Fig.~\ref{fig:cm} (f)).
This may be because the shallow fundal indentation of an arcuate uterus can resemble that of a small septate or T-shaped uterus on 3D US.
These overlapping morphological characteristics make it difficult for the model to learn discriminative features. Moreover, without explicit morphological constraints during training, the model may rely excessively on local appearance cues, thereby reducing generalizability and causing misclassifications, such as confusion with unicornuate uterus in the external test set (Fig.~\ref{fig:results_etest} (b)).
For the corresponding solutions, we believe that there are two feasible approaches. The first is to incorporate uterine morphology segmentation, as it could provide explicit anatomical guidance and encourage the model to focus on global uterine geometry. The second is to use controllable data synthesis to generate more diverse and balanced samples, thereby enriching morphological variations and potentially improving classification accuracy.

In future work, we will collect more underrepresented data (e.g., bicornuate uterus) to more comprehensively validate our method.
Additionally, we will develop stronger fine-grained feature discrimination strategies to better distinguish CUAs with similar characteristics.
Finally, we aim to promote broader multi-center collaboration and package the model as standalone software in clinical US systems to prospectively validate its clinical value and assist physicians in improving CUA diagnostic accuracy.
\appendix

\section*{Acknowledgments}
This work was supported by the Guangxi Key Research and Development Program (No. AB23026042), the National Natural Science Foundation of China (No. 12326619), the Frontier Technology Development Program of Jiangsu Province (No. BF2024078), and the Guangxi Natural Science Foundation (No. 2025GXNSFAA069471).

\bibliographystyle{elsarticle-num} 
\bibliography{ref}

\end{document}